\documentclass[accepted]{uai2021} % after acceptance, for a revised
\usepackage[american]{babel}
\usepackage[numbers, sort]{natbib}
\usepackage[normalem]{ulem} %for striking through during internal iterations and passes
\usepackage{mathtools} % amsmath with fixes and additions
\usepackage{booktabs} % commands to create good-looking tables
\usepackage{tikz} % nice language for creating drawings and diagrams
\usepackage{url}            % simple URL typesetting
\usepackage{amsfonts}       % blackboard math symbols
\usepackage{nicefrac}       % compact symbols for 1/2, etc.
\usepackage{microtype}      % microtypography
\usepackage{makecell}
\usepackage{graphicx}
\usepackage{adjustbox}
\usepackage{sidecap}
\usepackage{subcaption}
\usepackage{amsmath}
\usepackage{wrapfig}
\usepackage{lineno}
\usepackage{comment}
\usepackage{amssymb}
\usepackage{times}
\usepackage{epsfig}
\usepackage{parskip}
\title{Saliency Can Be All You Need \\In Contrastive Self-Supervised Learning}
\author[1]{\href{mailto:Veysel Kocaman <v.kocaman@liacs.leidenuniv.nl>?Subject=Your SVC-2002 paper}{Veysel Kocaman}{}} % Lead author
\author[2]{Ofer M. Shir}
\author[1]{Thomas Bäck}
\author[3]{Ahmed Nabil Belbachir}
 
\affil[1]{%
    LIACS\\
    Leiden University\\
    Leiden, The Netherlands
}
\affil[2]{%
    Computer Science Department\\
    Tel-Hai College and Migal Institute\\
    Upper Galilee, Israel
}
\affil[3]{%
    Norwegian Research Centre AS\\
    NORCE Technology\\
    Grimstad, Norway
}

\begin{document}
\maketitle

\begin{abstract}
We propose an augmentation policy for Contrastive Self-Supervised Learning (SSL) in the form of an already established Salient Image Segmentation technique entitled Global Contrast based Salient Region Detection.
This detection technique, which had been devised for unrelated Computer Vision tasks, was empirically observed to play the role of an augmentation facilitator within the SSL protocol.
This observation is rooted in our practical attempts to learn, by SSL-fashion, aerial imagery of solar panels, which exhibit challenging boundary patterns. 
Upon the successful integration of this technique on our problem domain, we formulated a generalized procedure and conducted a comprehensive, systematic performance assessment with various Contrastive SSL algorithms subject to standard augmentation techniques. This evaluation, which was conducted across multiple datasets, indicated that the proposed technique indeed contributes to SSL.
We hypothesize whether salient image segmentation may suffice as the only augmentation policy in Contrastive SSL when treating downstream segmentation tasks.
\end{abstract}

\section{Introduction}
\label{sec:introduction}

Despite recent advances in Computer Vision (CV), the effectiveness of visual recognition and learning still very much depends on manual annotations and on how well they represent the images at hand. 
Given the substantial amount of available unlabeled data and the costly manual annotation, new methods for online and unsupervised learning are crucial for achieving robust and efficient visual learning. 
Current unsupervised learning methods do not perform learning under a genuine unsupervised state -- they rather focus on transfer learning, or unsupervised tasks subject to strong features that were pre-trained in a supervised way. Therefore, there is a need to investigate algorithms for unsupervised learning of completely new categories, such as self-supervised learning (SSL) %[`study means' appeared in the original text ] and \os{study means?}\vk{can you clarify please ?} 
to learn in a continuous, long-term fashion. 
Indeed, SSL needs also to generalize the classical unsupervised learning scenario of physical objects to scenarios of higher-level actions and more complex activities, and at the same time develop capabilities to detect abnormalities in visual data ~\cite{leordeanu2020unsupervised},~\cite{chen2019self}.

The requirement for very large datasets of manually labeled instances may seem counter-intuitive since this is not how humans learn to recognize new objects. 
Humans are constantly fed with images through their eyes, and are able to learn an object's appearance and to distinguish it from other objects without knowing what the object exactly is. 
Moreover, collecting large-scale datasets is time-consuming and expensive, while the supervised approach to learn features from labeled data has almost reached its saturation due to the intense labor required in manually annotating millions of data instances. 
This is because most of the modern CV systems (that are supervised) try to learn some form of image representations by finding a pattern that links the data points to their respective annotations in large datasets ~\cite{jaiswal2020survey}. Moreover, the data annotation efforts vary from task to task, and it is estimated that the time spent on image segmentation and object detection (i.e., carefully drawing boundaries) is four times longer than the image classification itself~\cite{ciga2021learning}. The annotation efforts become significantly higher when it comes to highly regulated and specialized domains like medicine and finance in which the expertise level of the human annotator matters more than in any other domain. Moreover, supervised learning not only depends on expensive annotations but also suffers from other drawbacks such as generalization errors, spurious correlations, and being prone to adversarial attacks~\cite{liu2021self}.
%\tb{sounds like that is a formal fact? But probably better to say "significantly", instead?}\vk{fixed}

In this study, we explore the viability of using an unsupervised image segmentation technique called 'Global Contrast based Salient Region Detection (SGD)'~\cite{cheng2011niloy} as an augmentation policy (rather than a proxy task) in contrastive SSL methods and study its impact on downstream supervised image segmentation tasks in low data regimes. To the best of our knowledge, this is the first study that explores salient object segmentation techniques as an augmentation policy in contrastive SSL. 
%\vk{we may need a better intro for what we propose here. e.g. connecting the previous paragrahps to the essence of this study}

%Salient object detection (SOD) aims at highlighting visually salient object regions in images. Here, ‘visually salient’ describes the property of an object or a region to attract human observers’ attention~\cite{wang2021salient}. 

%In this study, we challenge this idea and claim that ~\textit{salient object detection algorithms can also be used as a proxy augmentation technique given the downstream task}. As the augmentations should make sense in terms of the downstream task, we will base our experiments on image segmentation and object detection tasks.

%Taking one step backward, we feel the urge to inform the reader about our ideation process. We came up with this idea while trying to find a way to segment the Photo-Voltaic (PV) Solar Arrays into smaller areas to improve the process of detecting and classifying the solar fault defects given the rough labels. Comprising of multiple solar panels and cells, PV arrays usually cover a large area and detecting.

We will show that using this SGD algorithm as an image augmentation policy in SSL produces better representations for downstream image segmentation tasks when compared to default augmentation policies commonly utilized in SSL methods.  
In order to make the integration of SGD into SSL pretraining routines feasible, we also devise a simple manipulation called offline augmentation with hashing that enables running comprehensive experiments with various parameters and configurations. We will also provide evidence that SGD-based augmentation policy in SSL performs better with low resolution images.  

The current study targets the following \textit{research questions}:
\begin{quote}
    Would SGD produce better representation when used as an augmentation policy in Contrastive SSL? Moreover, would salient image segmentation suffice as the only augmentation policy in Contrastive SSL when treating downstream segmentation tasks?
\end{quote}

The concrete contributions of this paper are the following:
%\vk{We may need to simplify these to give a clearer message}
%
\begin{itemize}
    \item Comparing a relatively old unsupervised image segmentation technique, Global Contrast based Salient Region Detection (SGD), with recent deep learning (DL)-based image segmentation algorithms.
    %state-of-the-art unsupervised deep learning (DL) based image segmentation algorithms.
    \item Evaluating the viability of a generative model (Pix2Pix) as a proxy for computationally expensive image augmentation methods.
    \item Devising an SGD-based efficient offline augmentation technique to incorporate any expensive augmentation policy in DL training routines.
    \item Employing SGD as an offline augmentation policy (rather than a proxy task) in contrastive SSL methods and study its impact on downstream supervised image segmentation and object detection tasks.
    %Derived from the previous item - removed to save space
    %\item Exploring the viability of replacing all of the default random augmentation policies with a single method (SGD) in contrastive SSL. %\tb{"augmentation"?}{\vk{fixed}}
    \item Illustrating that fine-grained details in high resolution images would negatively impact the performance of SSL when compared to the coarse-grained details in low resolution images.
    \item  Formulating a recommendation for choosing the most appropriate SSL method (accounting for the the augmentation technique, downstream task and even the imagery resolution).
\end{itemize}

The remainder of the paper is organized as follows: 
%Section~\ref{sec:relatedWork} explores  similar studies that aim to produce robust ConvNet backbones using various SSL techniques for downstream image segmentation and object detection tasks.
Section~\ref{sec:motivation+background} describes the concrete motivation that ignited this research, and then summarizes related work as well as various SSL approaches, including the role of data augmentation therein. It concludes with presenting the SGD method in detail. 
%\tb{sentence too long, one leses overview when reading, and its construction seems a bit odd.}\vk{simplified and rewritten}
Section~\ref{sec:ImplementationDetails} outlines the experimental setup regarding SGD, including the datasets the various preliminary efforts to make SGD-based augmentation policies viable in DL training routines. It then presents the attained results. 
%\tb{Likewise. Make them shorter.}\vk{simplified and rewritten}
Section~\ref{sec:discussion}  discusses the findings and proposes possible \textit{mechanistic} explanations, and finally Section~\ref{sec:conclusion} concludes by pointing out the key points and future directions.

% \td{MISSING: Problem Formulation, and especially Research Question -- \\
% ``can SGD effectively play the role of an augmentation facilitator?''
% }

% \vk{what about this:
% \textbf{whether SGD would produce better representation when used as an augmentation policy in contrastive SSL methods and  salient image segmentation may suffice as the only augmentation policy in Contrastive SSL when treating downstream segmentation tasks.}}
\section{Motivation and Background}\label{sec:motivation+background}
    \subsection{Related Work}
\label{sec:relatedWork}

%--- Academic siblings of our work, i.e. alternative attempts in literature at trying to solve the same problem ---
Semantic segmentation is the task of assigning each pixel to a specific class label. The class labels can be the same as for object detection, but unlike the object detection task, which labels each instance of an object as separate objects, semantic segmentation only assigns a pixel a specific class label and does not differentiate instances of objects. 

As the augmentations in any SSL method should be tailored \textit{a priori} and fit  in terms of the downstream tasks, devising an augmentation policy for downstream segmentation tasks depends on harnessing the intrinsic features that help model learn how to segment the objects in an image. 
This is usually achieved through auxilary tasks such as rotating, cropping, colorization etc. One of the most useful auxilary tasks can be regarded as image colorization that is introduced in ~\cite{zhang2016colorful} as a process of estimating RGB colors for grayscale images. The backbone network within the pretrained model performed well for downstream tasks like object classification, detection, and segmentation compared to other methods. 
The study in ~\cite{iizuka2016let} also suggested a similar technique to automatically colorize grayscale images by merging local information dependent on small image patches with global priors computed using the entire image. 
Since these two techniques employ a strategy of finding suitable reference images and transferring their color onto a target grayscale image, the semantic information plays a little role and has the potential to actually hamper the SSL. 
Probably one of the most useful colorization tasks is suggested by ~\cite{larsson2016learning}, in which a system must interpret the semantic composition of the scene (what is in the image) as well as to localize objects (where things are) to incorporate semantic parsing and localization into a colorization system. In a generative manner, ~\cite{pathak2016context} proposed image inpainting, a context-based pixel prediction where the network understands the context of the entire image as well as the hypothesis for missing parts, hence assisting in extracting the semantic information.

There are many other auxiliary tasks proposed in SSL but most of these methods focus on basic inherent visual features, like image patches and rotation, which are very simple tasks that are not likely to completely learn the semantics and the spatial features of the image. Actually, the default augmentation policies in contrastive SSL methods are the derivation of these auxiliary tasks. To the best of our knowledge, salient image segmentation has not been used as an auxiliary task in any SSL method before, let alone contrastive SSL.

%Academic siblings of our work, i.e. alternative attempts in literature at trying to solve the same problem.  Goal is to “Compare and contrast” - how does their approach differ in either assumptions or method? If their method is applicable to our problem setting I expect a comparison in the experimental section. If not there needs to be a clear statement why a given method is not applicable. 
%Note: Just describing what another paper is doing is not enough. We need to compare and contrast.

    \subsection{Concrete Background}
\label{sec:background}

To avoid time-consuming and expensive data annotations, as an alternative to data-hungry supervised learning methods, many SSL methods were proposed to learn visual features from large-scale unlabeled images or videos without using any human annotations. Formally, SSL is a subset of unsupervised learning methods, referring to learning methods in which ConvNets are explicitly trained with automatically generated labels; so the supervision comes from structure of the data itself. Since no human annotations are needed to generate pseudo labels during self-supervised training, very large-scale datasets can be used for SSL training. 
Trained with these pseudo labels, self-supervised methods typically achieve promising results, while gradually closing the performance gap with respect to supervised methods~\cite{jaiswal2020survey}. 

On the other hand, in order to address the lack of annotated datasets or dataset shifts, there are various efforts towards developing zero-shot or few-shot learners. Zero-shot learners (ZSL) aim to predict the correct class without being exposed to any instances belonging to that class in the training dataset, while few-shot learners (FSL) attempt to accomplish the same when a small number of examples are available in the training dataset. In short, ZSL/FSL and SSL have important commonalities in that they can be applied in situations where annotated data is scarce. 
This close relationship allows some of the SSL methods to gain semantic scene understanding through a pretraining process. For example, certain attention heads in DINO~\cite{caron2021emerging} are found to be discovering and segmenting objects in an image or a video with no supervision and without being given a segmentation-targeted objective. However, it is a computationally expensive process to segment an image using transformer based architectures and usually this does not work well.

Another related concept, entitled semi-supervised learning, refers to a learning problem involving a small portion of labeled examples and a large number of unlabeled examples from which a model must learn and make predictions on new examples. It is halfway between supervised and unsupervised learning, and directly relevant to a multitude of practical problems where it is relatively expensive to produce labeled data.  Some of the most popular semi-supervised learning methods include pseudo learning~\cite{lee2013pseudo} (a model is trained on a labeled dataset and used to predict pseudo-labels for the unlabeled data) and a noisy student ~\cite{xie2019self} (training two separate models called ``Teacher'' and ``Student''). Being central to our experiments, we will focus more on SSL and SGD in this section.

%\tb{what do you mean by "in a parallel"?} \vk{'in a parallel with similar needs regarding lack of annotated datasets' is replaced by 'in order to deal with the lack of annotated datasets'}
%\tb{labels are scarce, not data?}\vk{fixed}
%\tb{"this does not"?}\vk{fixed} this does not work well.

%By discovering object parts and shared characteristics across images, the model learns a feature space that exhibits a very interesting structure. If we embed ImageNet classes using the features computed using DINO, we see that they organize in an interpretable way, with similar categories landing near one another. This suggests that the model managed to connect categories based on visual properties, a bit like humans 

\subsubsection{Self-Supervised Learning (SSL)}
%Academic Ancestors of our work, i.e. all concepts and prior work that are required for understanding our method. Includes a subsection Problem Setting which formally introduces the problem setting and notation (Formalism) for our method. Highlights any specific assumptions that are made that are unusual. Self-supervised learning is a machine learning process where the model trains itself to learn one part of the input from another part of the input. It is also known as predictive or pretext learning. 
Explicitly, SSL is a machine learning process where the model trains itself to learn one part of the input from another part of the input. In other words, the model learns from labels that are presumably already intrinsic in the data itself. This process is also known as predictive or pretext learning and the unsupervised problem is transformed into a supervised problem by auto-generating the labels. 
To effectively exploit the huge quantity of unlabeled data, it is crucial to set the right learning objectives, in order to obtain the appropriate supervision from the data itself. 

%Transfer learning (TL) has been presented as an effective solution for constructing robust feature representations when the training set for a given problem is small. As its name suggests, TL aims to transfer knowledge and learned features from one task (the source task) to another related target task, just as a person can utilize the same knowledge across different projects. To do this, TL trains the model on a large labeled dataset and then treats this model as a starting point in the target task’s training, without learning from scratch. This dataset creates the target task’s representation model, using the same architecture as the source task. The initialized representation network in the target task is then further trained on the target dataset. So we can say that the workflows of SSL and TL are similar, with only slight differences. The key difference between TL and SSL is that TL pre-trains on labeled data, whereas SSL utilizes unlabeled to learn features and need only a small number of labelled example~\cite{albelwi2022survey}.

Within the context of utilizing unlabeled data to learn the underlying representations, SSL can be organized as (i) handcrafted pretext tasks-based, (ii) contrastive learning-based and (iii) clustering learning-based approaches. In handcrafted pretext tasks-based method, a popular approach has been to propose various pretext tasks that help in learning features using pseudo-labels while the networks can be trained by learning objective functions of the pretext tasks and the features are learned through this process~\cite{jaiswal2020survey}. Tasks such as image-inpainting~\cite{pathak2016context}, colorizing gray-scale images~\cite{zhang2016colorful}, solving jigsaw puzzles~\cite{noroozi2016unsupervised}, image super-resolution~\cite{ledig2017photo}, video frame prediction~\cite{xu2019self}, audio-visual correspondence~\cite{afouras2021self}, to mention the most prominent, have proven to be effective for learning good representations. In doing so, the model learns quality representations of the samples and is used later for transferring knowledge to downstream tasks. 
The selection of an appropriate proxy task (pretext) is critical to the effectiveness of self-supervised learning. It requires careful design, and indeed, numerous researchers investigated various approaches for given downstream tasks. For example,~\cite{dhere2021self} proposed a proxy task to classify whether a given pair of kidneys belong to the same side; with the assumption that the network needs to develop an understanding of the structure and sizes of the kidneys.

\begin{figure}[hbt]
  \centering
  \includegraphics[width=1.0\columnwidth]{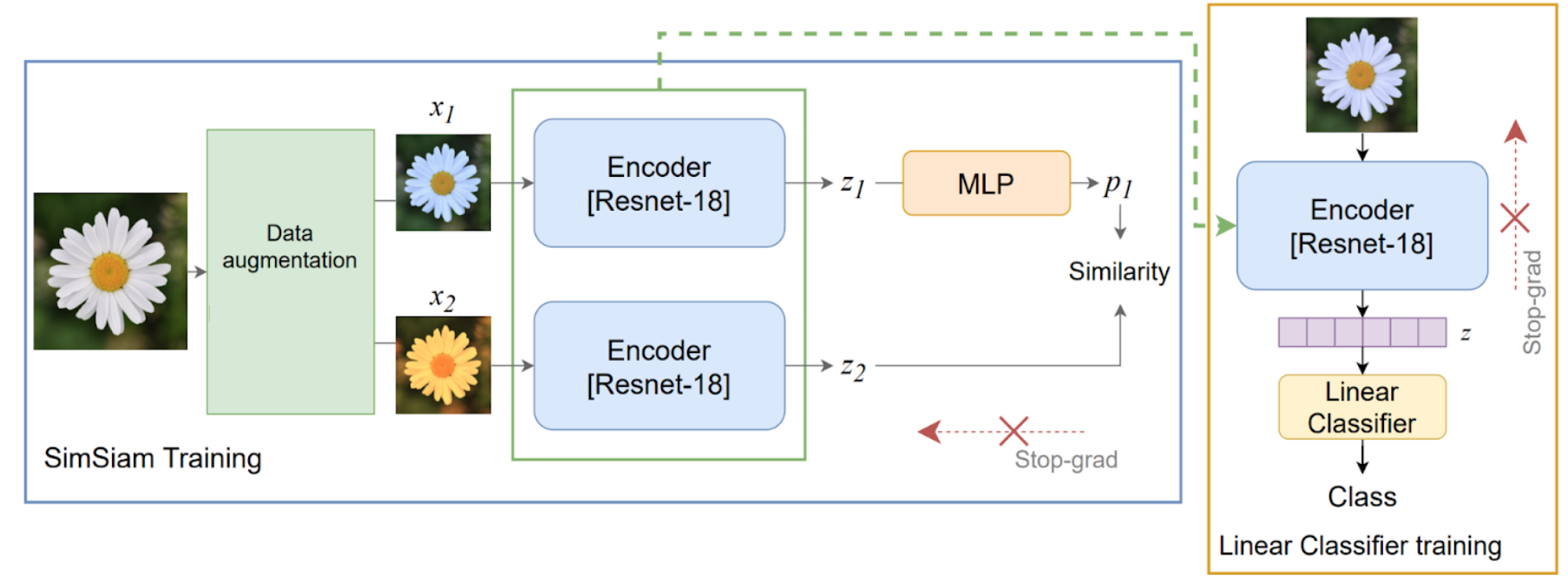}
\caption{Linear evaluation protocol to evaluate the learned representations of a pretraining process of SimSiam~\cite{nakamura2022self}.}
\vspace{-5mm}
\label{figure:simsiam_linear_clf}
\end{figure}

On the other hand, contrastive learning (CL) is a training method wherein a classifier distinguishes between “similar” (positive) and “dissimilar” (negative) input pairs. 
It is essentially the task of grouping similar samples closer to each other, unlike setting diverse samples far from each other. 
During training, the augmented version of the original sample is considered as a positive sample, and the rest of the samples in the batch/dataset (depends on the method being used) are considered negative samples. Next, the model is trained in a way that it learns to differentiate positive samples from the negative ones. Constructing positive and negative pairs via data augmentation allows the model to learn from inter-class variance (uniformity) by pushing negative pairs far away, and from intra-class similarity (alignment) by pulling positive pairs together. %(see Figure ~\ref{figure:simclr} for an illustration).
%~\vk{add a sentence for 3rd approach of SSL --> clustering learning-based}
%\begin{figure}[h]
  %\centering
  %\includegraphics[width=0.8\columnwidth]{figures/SimCLR.png}
%\caption{Contrastive learning methods train a model to cluster an image and its slightly augmented version in latent space, while the distance to other images should be maximized (taken from the official SimCLR repo- https://simclr.github.io/.)}
%\vspace{-5mm}
%\label{figure:simclr}
%\end{figure}
The core concept is to maximize the dot product between the feature vectors which are similar and minimize the dot product between those of which are not similar. CL methods use contrastive loss that is evaluated based on the feature representations of the images extracted from an encoder network. The popular methods that recently started to produce results comparable to the state-of-the-art supervised learning methods, even with less labelled images, can be regarded as DINO~\cite{caron2021emerging}, SwAV~\cite{caron2020unsupervised}, MoCo~\cite{he2020momentum}, \cite{chen2020improved} , BYOL~\cite{grill2020bootstrap}, SimSiam~\cite{nakamura2022self} and SimCLR, \cite{chen2020big}, \cite{chen2020simple}. The representation extraction strategies differ from one method to another (e.g. BYOL does not even need negative pairs) but the changes are very subtle and without rigorous ablations, it is hard to tell which one works better on a case at hand. The widely used approach to evaluate the learned representations through the SSL's pre-training process is the linear evaluation protocol~\cite{kolesnikov2019revisiting}, where a linear classifier (e.g., SVM, Logistic Regression, etc.) is trained on top of the frozen backbone network (image representations are derived from the final or penultimate layers of the backbone network as a feature vector). Finally, the test accuracy is used as a proxy for representation quality (Figure~\ref{figure:simsiam_linear_clf}). 
%\os{\sout{in} per a feature vector(?)}\vk{fixed: as a feature vector.}

Data augmentation plays an important role in the learning process of contrastive SSL methods, while the composition of multiple data augmentation operations is crucial in defining the contrastive prediction tasks that accomplish effective representations. In addition, contrastive SSL benefits from stronger data augmentation, when compared to supervised learning. 
%\os{UNCLEAR: That colour distortion and cropping are the key transformations to produce our views for the considered dataset~\cite{chen2020simple}. }
Augmentations can be regarded as an indirect way to pass human prior knowledge into the model, and an effective augmentation should discard the unimportant features for the downstream task (i.e., removing the ``noise'' to classify an image). 
The nature of the selected augmentations should fit the downstream task, maintain the image semantics (meaning), introduce the model with challenging assignments/tests. 
Their selection depends upon the dataset's underlying distribution and cardinality, whereas a recent study~\cite{balestriero2022effects} further claims that augmentations are problem-class-dependent. 
%\os{ UNCLEAR: give the model a hard time and are dependent on the dataset’s diversity and size.} \vk{adding more clarity: in contrastive SSL, we feed pairs/different views of the same image to the network and these views should be visually dissimilar in a way that the model should have a hard time while bringing them together if the views are coming from the same image or pushing them apart if they are coming from different images. If a weak augmentation is applied, it will be an easy job for the network to puss/pull; hence the learning will not be strong enough. Dataset’s diversity and size play a role here as the more diverse the dataset, the easier it gets for the model to push dissimilar views (in low diversity datasets in which there many similar images, it is harder to push dissimilar views)}

The most popular data augmentation techniques in contrastive SSL include rotation, crop, cut, flip, color jitter, blur, Gaussian noise and gray scale. Almost all of the contrastive SSL methods use these (or similar) augmentation techniques at certain degrees no matter what the downstream task is. Selection of the most suitable augmentation policy is a crucial step in contrastive SSL. \cite{chen2020simple} systematically studied the impact of data augmentation and observed that no single transformation suffices to learn good representations, even though the model can almost perfectly identify the positive pairs in the contrastive task. When composing augmentations, the contrastive prediction task becomes harder, but the quality of representation improves dramatically. Compared to handcrafted pretext tasks-based methods, contrastive SSL methods are easier to implement but the pretraining process is computationally expensive as they require large batches and large amount of unlabelled datasets. Nevertheless, the pretrained backbone ConvNets through contrastive SSL methods generally produce better latent representations and perform well on downstream tasks.

Importantly, even though classical image segmentation methods can be regarded as one of the pretext tasks in SSL, the downstream segmentation and object detection tasks usually try to detect/segment certain salient objects and areas in an image, rather than entire textures, which unsupervised segmentation algorithms generate. 
While essentially solving a segmentation problem, salient object detection and segmentation approaches segment only the salient foreground object from the background, rather than partitioning an image into regions of coherent properties as in general segmentation algorithms. That is, using a salient image segmentation in SSL is likely to generate better representations when the downstream task is defined as either image segmentation or object detection.

\subsubsection{Global Contrast based Salient Region Detection (SGD)}

%Being able to automatically, efﬁciently, andaccurately estimate salient object regions, however, ishighly desirable given the immediate ability to charac-terise the spatial support for feature extraction, isolatethe object from potentially confusing background, andpreferentially allocate ﬁnite computational resources forsubsequent image processing

SGD, also known as SaliencyCut~\cite{cheng2011niloy}, is an automated unsupervised salient region extraction method, an improved iterative version of GrabCut~\cite{rother2004grabcut}, which introduced a contrast analysis method to integrate spatial relationships into region-level contrast computation. 
In GrabCut, the user initially draws a rectangle around the foreground region in an image, and then the algorithm iteratively segments it to get the best result. 
The executed steps are (i) estimating the color distribution of the foreground and background via a Gaussian Mixture Model (GMM), (ii) constructing a Markov random field over the pixels labels (i.e., foreground vs. background), and (iii) applying a graph cut optimization to arrive at the final segmentation. Instead of manually selecting this rectangular region to initialize the process, SaliencyCut is using histogram-based contrast (HC) and region-based contrast (RC) techniques (that are also suggested in the same paper) to create saliency maps and then binarizes this map using a fixed threshold (e.g., value of $70$). 

In RC, the input image is ﬁrst segmented into regions, then the color contrast at the region level is computed, and saliency for each region is ﬁnally deﬁned as the weighted sum of the region’s contrasts to all other regions in the image. The weights are set according to the spatial distances with farther regions being assigned smaller weights. 

In HC, the number of colors needed to consider is reduced to 1728 by quantizing each color channel to have 12 different values (12 color values per channel in R-G-B). Considering that color in a natural image typically covers only a small portion of the full color space, the number of colors is further reduced by ignoring less frequently occurring colors. 
By choosing more frequently occurring colors, and by ensuring that they cover the colors used \textit{de facto} by more than 95\% of the image pixels, we are typically left with around n = 85 colors. 
The colors of the remaining pixels, which comprise fewer than 5\% of the image pixels, are replaced by the closest colors in the histogram.

\begin{figure}[h]
\includegraphics[width=1.0\columnwidth]{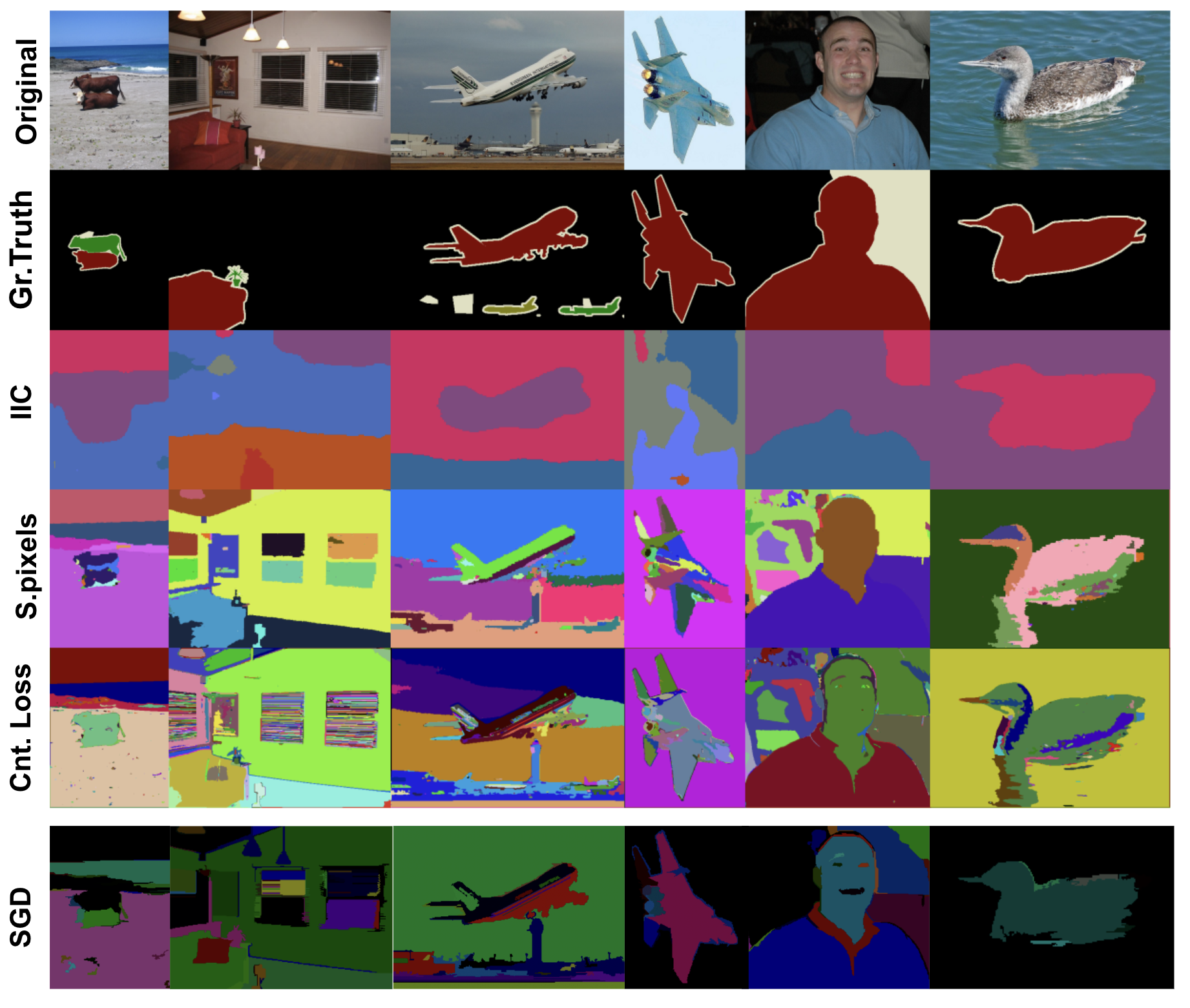}
\caption{Comparing the outcomes of unsupervised versus SGD-based segmentation on PASCAL VOC 2012.  Invariant Information Clustering (IIC)~\cite{ji2019invariant}, Superpixels~\cite{kanezaki2018unsupervised}, Continuity Loss~\cite{kim2020unsupervised}. Different segments are shown in different colors (the images in the first 5 rows are taken from~\cite{kim2020unsupervised}). 
Notably, SGD was developed at least 8 years prior to IIC and Superpixels, and can still perform comparatively better on some cases.}
\label{figure:segm_tile}
\end{figure}

Once the saliency map is initialized with RC and HC, GrabCut is iteratively run (i.e., iterative refinements) to improve the SaliencyCut result. 
After each iteration, dilation and erosion operations are used on the current segmentation result to get a new trimap for the next GrabCut iteration. 
During this iterative refinement, adaptive ﬁtting is used (regions closer to an initial salient object region are more likely to be part of that salient object than far-away regions). Thus, the new initialization enables GrabCut to include nearby salient regions, and exclude non-salient regions according to color dissimilarity. 
See~\cite{cheng2011niloy} for more details regarding SGD. 
Finally, Figure~\ref{figure:segm_tile} presents the outcomes' comparison of SGD-based segmentation versus latest unsupervised segmentation on PASCAL VOC 2012~\cite{everingham2010pascal}.

%\tb{Impossible to read the row labels in the figure; maybe explain explicitly in the caption what the rows are.}\vk{fixed -- row labels are scaled up}
 
%\subsection{Problem Setting}

 %-- introducing the problem setting and notation for our method. Highlights any specific assumptions that are made that are unusual. ---

\section{Implementation, Setup \& Results}\label{sec:ImplementationDetails}

Given the aforementioned research question, we derive concrete experimentation tasks:

\begin{enumerate}[noitemsep, topsep=0pt]
    \item Preliminary: Applying SGD to the PVs' datasets
    \item Obtaining a solid and an efficient implementation
    \item Testing the primary hypothesis: SGD as an augmentation policy
\end{enumerate}
%\subsection{`Experimental Planning}

\subsection{Setup and Datasets}
We used the following datasets in the current study:
\begin{itemize}
    \item \textbf{Aerial Drone Images for Solar Photovoltaic (PV) Panels Defects (NORCE-PV):} This dataset is provided by NORCE (Norwegian Research Centre) and has 790 high resolution aerial drone images. It is originally gathered for PV panels defect detection classification and annotated by NORCE internally. In this study, we will not be dealing with defect classes but the images themselves to run the experiments. The dataset is not publicly available.
    \item \textbf{Multi-resolution PV Dataset From Satellite and Aerial Imagery (MultiRes-PV):} This dataset includes three groups of PV samples collected at the spatial resolution of 0.8m, 0.3m and 0.1m, namely PV08 from Gaofen-2 and Beijing-2 imagery, PV03 from aerial photography, and PV01 from UAV orthophotos. In this study, we only used PV01 rooftop images (645 in total) that comes with segmentation masks that come within 256x256 size~\cite{jiang2021multi}. 
    \item \textbf{CIFAR10:} The CIFAR-10 dataset consists of 60000 32x32 colour images in 10 classes, with 6000 images per class. There are 50000 training images and 10000 test images~\cite{krizhevsky2009learning}.
    \item \textbf{STL10:} The STL-10 dataset is a subset of ImageNet, consists of 1300 labelled, 100000 96x96 unlabelled colour images in 10 classes. In particular, each class has fewer labeled training examples than in CIFAR-10, but a very large set of unlabeled examples is provided to learn image models prior to supervised training~\cite{coates2011analysis}.
\end{itemize}

\subsection{Preliminary: Running SGD on NORCE-PV and MultiRes-PV Datasets}

\begin{figure}[h]
  \centering
  \includegraphics[width=1.0\columnwidth]{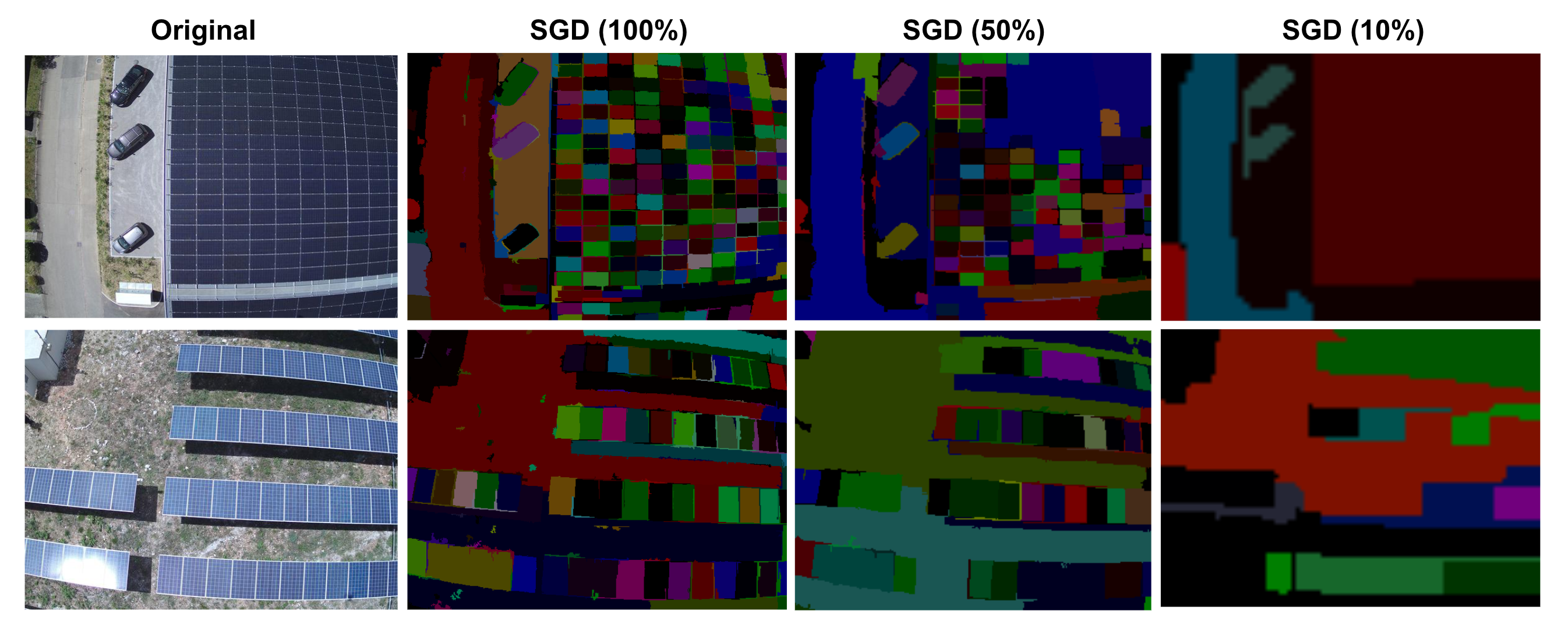}
\caption{Segmentation produced by SGD from NORCE-PV images with various sizes.}
\label{figure:sgd_solar_norce}
\end{figure}

\begin{figure}[hbt]
  \centering
  \includegraphics[width=1.0\columnwidth]{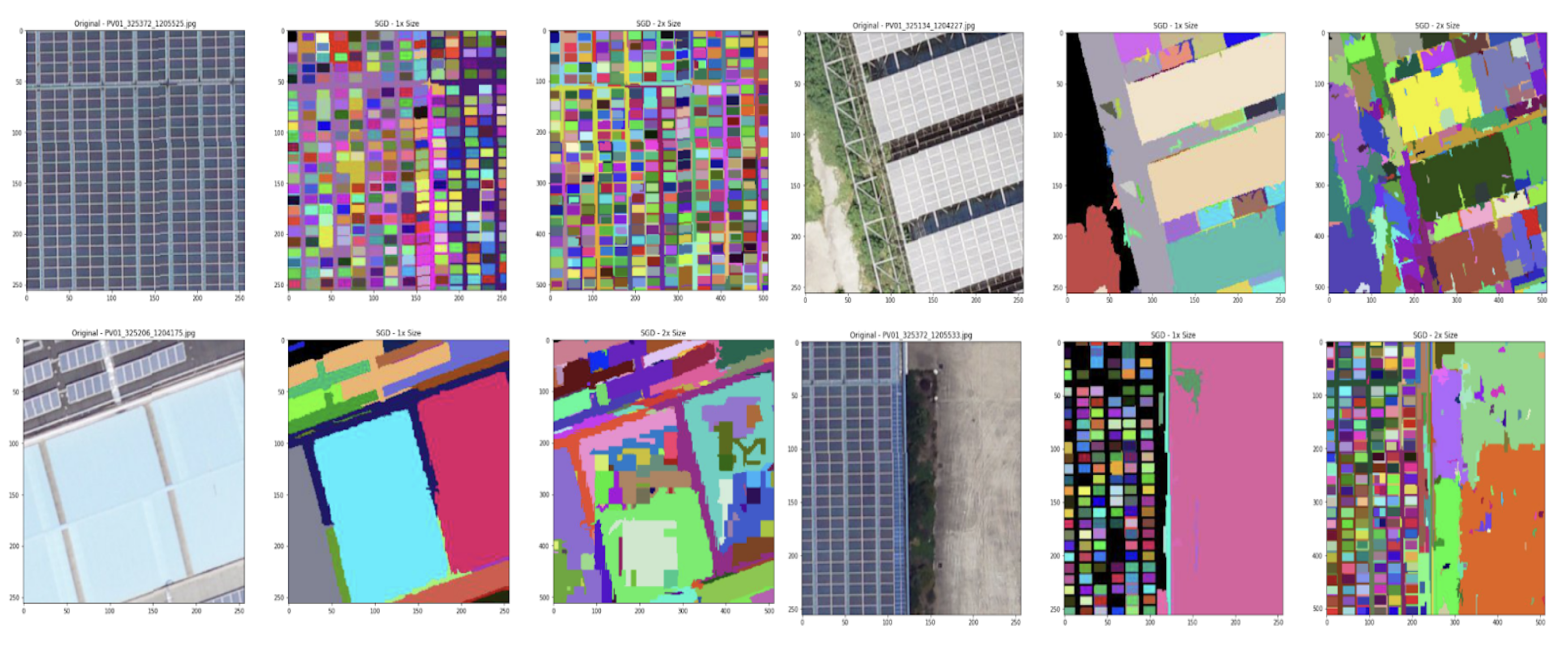}
\caption{Segmentation produced by SGD from low resolution MultiRes-PV images with original versus 200\% rescaled.}
\label{figure:SGD_triples_resized}
\end{figure}

SGD is a computationally expensive process and it produces segmentations on various level of details given the size of an image. For instance, it takes $\sim 2$min to generate a saliency map (segmentation) of an 600x450 RGB image, while it takes around $\sim 1$sec to accomplish the same process on 60x45 (10\% of the original image). 
In the cases wherein a highly detailed segmentation map is needed, SGD implementation would be practically impossible to implement on real time. In some other cases wherein a rough segmentation map would suffice, applying SGD on a reduced size of an image would still produce useful segmentation. On the other hand, when we apply SGD on low resolution images like MultiRes-PV images (256x256), we may need to upscale the image to get a better segmentation. 
The outcome of applying SGD on various sizes of NORCE-PV (high resolution) and MultiRes-PV (low resolution) images is presented in Figures~\ref{figure:sgd_solar_norce} and~\ref{figure:SGD_triples_resized}, respectively. 

\begin{figure}[hbt]
  \centering
  \includegraphics[width=1.0\columnwidth]{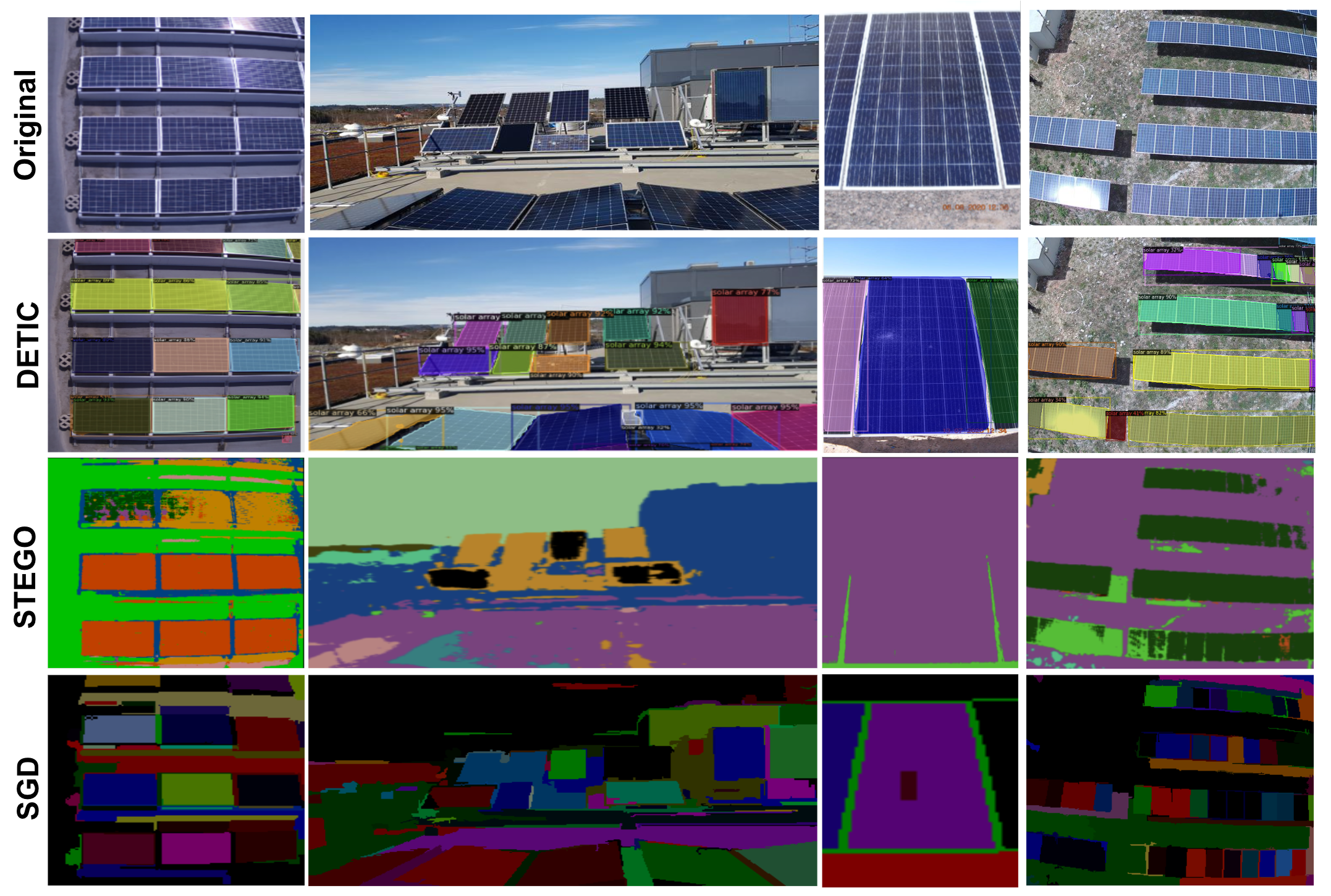}
\caption{Comparison of transformer-based zero-shot segmentation methods versus SGD on NORCE PV dataset.}
\label{figure:norce_sgd_vs_other_segm}
\end{figure}

We also run STEGO~\cite{hamilton2022unsupervised} and DETIC~\cite{zhou2022detecting}, two recent popular zero-shot DL-based image segmentation algorithms, on our datasets and compare results with SGD. 
As evident in Figure~\ref{figure:norce_sgd_vs_other_segm}, DETIC obtains impressive results as it is already capable of detecting solar arrays from natural images (it is trained to detect 20 thousand different objects out of the box). Compared to the heavy STEGO algorithm, developed just a few months ago, SGD performs comparatively better.

%\subsection{Speed Optimization of SGD for DL Training Routine}
\subsection{An Efficient Implementation}
Employing SGD on any DL training routine as an augmentation or transformation policy is highly inefficient in terms of computational costs. Notably, using GPU is not an option due to a layered (a mixture of OpenCV and Python functions) image preprocessing techniques along with RC and HC methods. The fact that thousands of images are to pass through this process in each epoch renders the integration of SGD in the DL training process practically infeasible. 
Our tests indicate that training a Resnet-18 ConvNet with SGD augmentation to classify PV defects using the NORCE-PV dataset (790 images, shape 32x32), for one epoch, takes $\sim 50$min on Tesla K80 (11.5GB) GPU.

\subsubsection{Image to Segmentation Map Translation via Pix2Pix}
\begin{figure}[!htb]
  \centering
  \includegraphics[width=1.0\columnwidth]{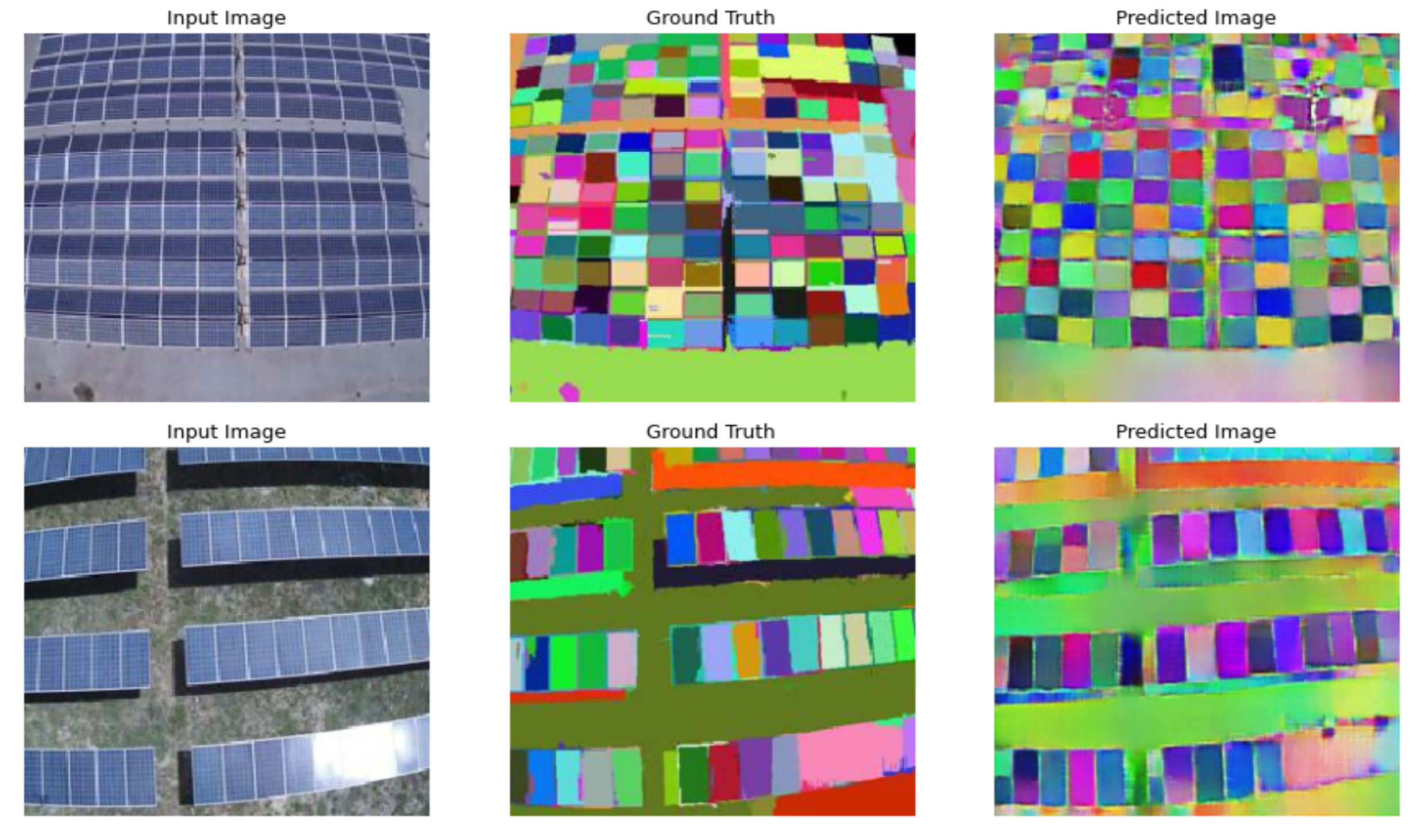}
\caption{The second column of this figure shows the SGD segmentation of the original images and the images generated with Pix2Pix image translation model are shown at the third column. }%Even though the Pix2Pix model managed to generate meaningful SGD representations from raw images, the information loss and distortion were clear and the speed gain was insufficient.}
%Since SGD is a computationally expensive process and cannot be used as an contrastive SSL augmentation technique, we tried to train
%\tb{tried to train - so what is the value of this figure? Focus on that!} \vk{rewritten}

\label{figure:pix2pix}
\end{figure}

To achieve feasiblity, we firstly thought about training a generative DL model to learn SGD segmentations via image translation and then replace the SGD process with this generative model. 
We chose the Pix2Pix architecture~\cite{isola2017image} for this purpose and trained several models. We basically fed the original images and SGD-segmented version to the network and trained further to get the model learn translating one image to another. Using this Pix2Pix model within the Resnet-18 training routine to replace SGD reduced the duration from $\sim 50$min to $\sim 30$min per epoch. 
Even though the results look promising (Figure~\ref{figure:pix2pix}), there are several drawbacks, e.g., information loss during the translation process, being outperformed on image instances that were under-represented, and the limited speed-improvement gains -- which altogether render this process unviable in practice.
%\os{UNCLEAR: low speed improvement gained} \vk{clarity: the speed improvement gained is still not big enough to use Pix2Pix in a DL routine as we cannot afford any augmentation that takes more than 0.5 sec} make this process still not viable for practical experimentation. 

\subsubsection{Offline Augmentation with Hashing}
%\tb{you are telling this a bit like a story - we did this, then this, etc. For a scientific paper, it will be better to "sell" this as a comparison of methods, or to drop the not so useful first steps and simply provide the best method that was found, compared to SOTA.}\vk{rewritten}
SGD is a computationally expensive method that results in a speed bottleneck when used in DL training routines and it might be one of the reasons that such a strong segmentation method could not establish itself despite all the latest advancements in similar fields. 
%In order to integrate SGD into the training and iterate faster with various hyperparameters, we experimented with various techniques %Then we came up with a simple trick to remove all the barriers regarding this speed issue: Offline augmentation by hashing. 
Since SGD is basically a deterministic approach, the segmentation map is always the same except colors; hence the contours and overall shapes are always identical for an image. 
In other words, even if the pixel colors change at each iteration of SGD, the overall shapes and lines in an image are always the same, unless resizing is applied. Experimenting around this attribute, we devised an unprecedented solution by running SGD once for all the images in the dataset and creating a segmentation mask for each image and then reusing that every epoch during training without running SGD again and again. In order to simulate the random colors and make the model invariant to colors (so it can focus more on other features, e.g., contrast, shapes, lines etc.), we also applied random color jittering and swapping from the same color palette utilized by SGD. 
Next, we describe the explicit steps:%~\vk{add pseudo code block with an example and also add some sample images after jittering}

\begin{enumerate}[(Step-1)]
    \item Read every image in the dataset as a \texttt{numpy} array and hash it to store the entire image as a single hash code (string),
    \item Run SGD over every image in the dataset and save the segmentation map as an image to the disk,
    \item Create a dictionary (key-value pairs) with the hash strings (of Step-1) as keys and the file path of every segmentation map as the associated value (if reading from disk becomes a bottleneck, all of the segmentation maps can be read at once and stored in the memory as an array),
    \item In model training with original images, get the hashmap of every image array and find the file path of corresponding segmentation map from the dictionary above at each iteration, and read that image from the disk or from memory if it is stored in memory, 
    \item Apply color jittering to randomly swap the colors of the augmented image and use that as a proxy augmented image during the rest of the process.
\end{enumerate}

%This simple trick reduced the time for applying the SGD process to practically zero seconds as the segmentation maps are read from disk/memory at real time with no cost.
This technique allowed SGD process to run instantly (practically below $1$sec) as the segmentation maps are read from disk/memory at real-time at no cost.
%\tb{If it is such a simple trick, is it worth describing it? Of course I think yes, but then present it as a generic method that can be used to speed up SGD.}\vk{you mean adding more details and some pseudo code to illustrate the steps ?}

\subsection{Using SGD as an Augmentation Policy in Contrastive SSL Algorithms}

In the Contrastive SSL pre-training routine, the augmented version of the original sample is considered as a positive sample, and the rest of the samples in the batch/dataset (depends on the method being used) are considered negative samples. Then, the model is trained in a way that it learns to differentiate positive samples from the negative ones. In doing so, the model learns quality representations of the samples and is used later for transferring knowledge to downstream tasks.

A predefined CV architecture (e.g., ResNet50, VGG19, EfficientNet etc.) is typically used as a ConvNet backbone, and the weights (model parameters) are updated during this contrastive SSL process. 
Then, the backbone is saved and used in another architecture as a starting point or just as a facilitator of feature extraction (representations from one of the last fully connected (FC) layers).

In this study, in order to quickly iterate across various heavy augmentation combinations under limited computational resources, we used a lightweight ResNet18 architecture and then pretrained several backbones with the combination of SGD versus standard image augmentation techniques (crop, grayscaling, jittering etc.) using the following SSL algorithms: SimSiam~\cite{nakamura2022self}, BYOL~\cite{grill2020bootstrap}, SimCLR~\cite{chen2020simple}, MoCo~\cite{he2020momentum}, SwAV~\cite{caron2020unsupervised} and Barlow Twins~\cite{zbontar2021barlow}. 
Then we tested the effectiveness of these backbones in downstream image clustering, image segmentation, object detection and classification problems.

\subsubsection{Image Clustering Using SSL Backbones}
\begin{figure}[hbt]
  \centering
  \includegraphics[width=1.0\columnwidth]{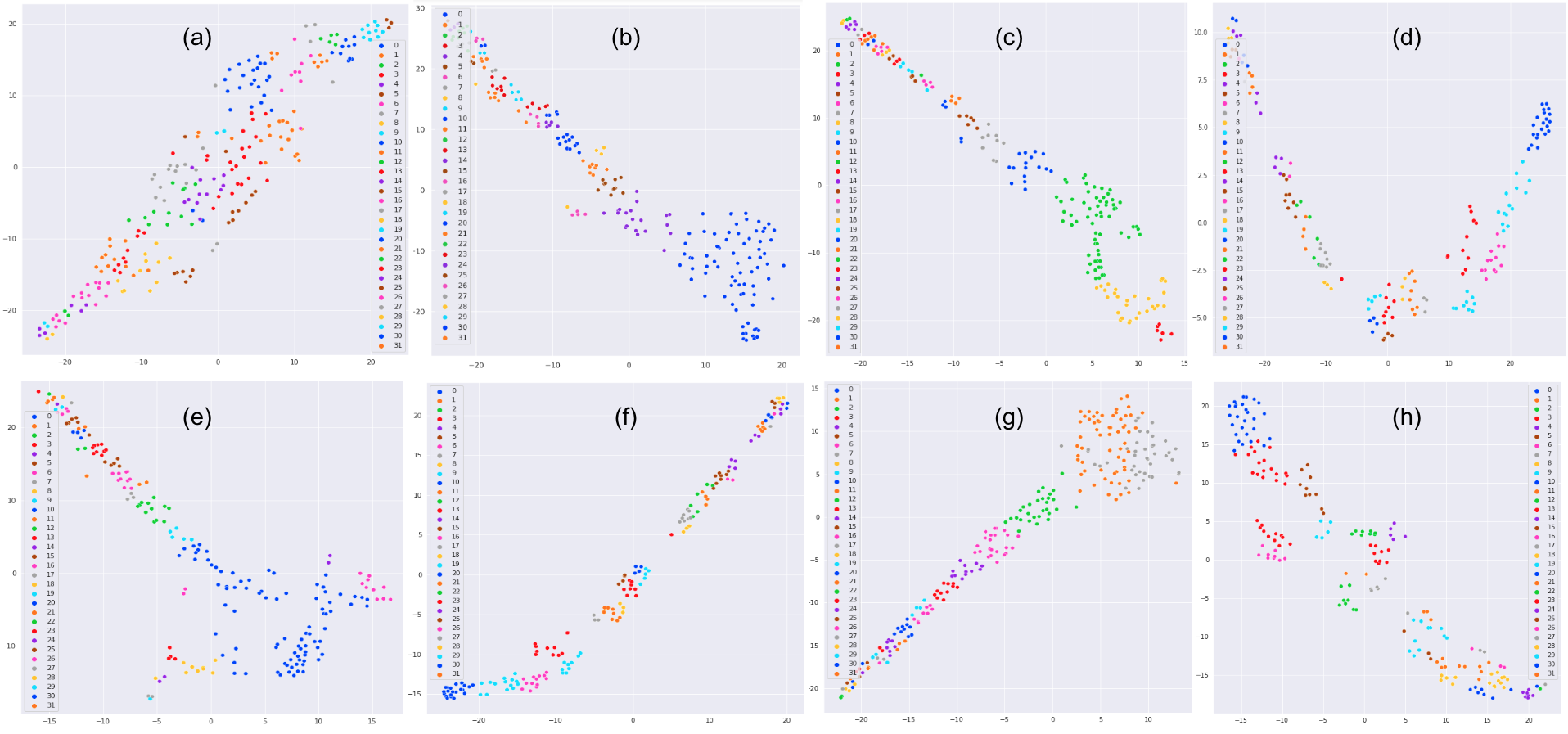}
\caption{Clustering partial NORCE-PV dataset image representations produced by (a) ResNet18 ImageNet weights (b) SimSiam random aug.~(c) SimSiam random + PixPix based SGD aug.~(d) SimCLR random + SGD aug.~(e) SimSiam random + SGD aug.~f) SimSiam with PixPix based SGD aug.~(g) SwAV with SGD aug. ~(h) SimSiam with SGD aug. }
\vspace{-5mm}
\label{figure:cluster_charts_SGD}
\end{figure}

Using the 80\% of the entire NORCE-PV dataset with no label, we trained SvAW, SimSiam and SimCLR models for 100 epochs with the combinations of default SSL augmentations and SGD (with offline augmentation through hashing as well as Pix2Pix versions), used their backbone networks to extract the image representations (vectors) of 20\% of the dataset, and then used KMeans and t-SNE methods to cluster and visualize the clusters. 
The outcome is presented in Figure~\ref{figure:cluster_charts_SGD}. 
Evidently, SGD, when pretrained on unlabelled images, enables better image representations (vectors, embeddings) in all of the tested cases. 
The more distant the clusters, the better the representations generated by the SSL backbone network. 
In fact, observing this clear delineation between the clusters on a 2D space had originally given us an idea of applying SGD as an augmentation on larger scales with other combinations on a downstream image segmentation task.
%\tb{Figure is hard to read, but could be ok.}

\subsubsection{Image Segmentation and Object Detection Using SSL Backbones}
\begin{figure}[hbt]
  \centering
  \includegraphics[width=1.0\columnwidth]{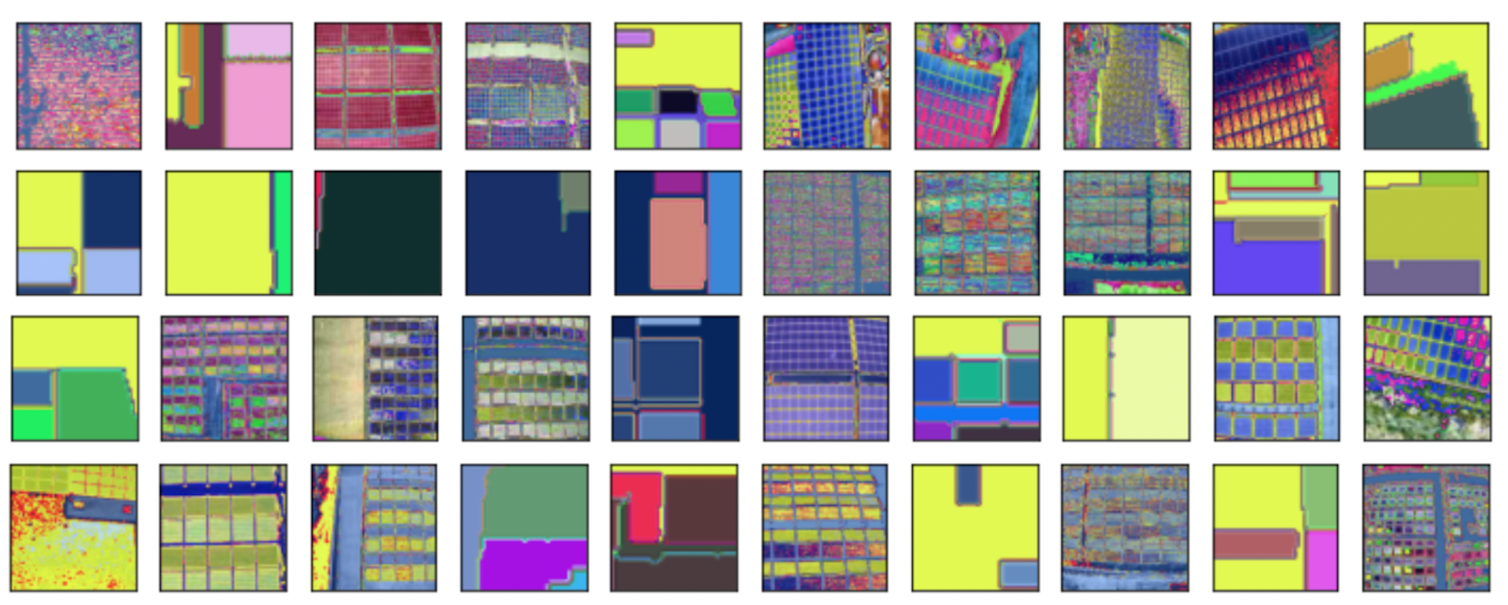}
\caption{This figure shows a batch of 40 NORCE-PV images. Every image in this batch is segmented by SGD at first and then the other default random augmentations are applied. Basically, this is what the model sees at each iteration.}
\vspace{-5mm}
\label{figure:SSL_aug_SGD_batch}
\end{figure}

As a downstream image segmentation architecture, we used Detectron2~\cite{wu2019detectron2} framework on MultiRes-PV dataset to detect and segment solar panels from rooftop aerial images. In order to pretrain the SSL algorithms, the following augmentation policies are applied ($p$ within the parenthesis corresponds to the probability of applying the respective technique, and sample augmentations are shown in Figure~\ref{figure:SSL_aug_SGD_batch}): 
%
%~\textit{Default SSL augmentations, SGD (p=1.0), SGD (p=1.0) with default SSL augmentations, SGD (p=0.5) with default SSL augmentations, SGD (p=1.0) with jittering (p=1.0), SGD (p=0.5) with jittering (p=1.0), SGD (p=0.8) with jittering (p=0.8)}. Sample augmentations can be seen at Figure ~\ref{figure:SSL_aug_SGD_batch}.
%
\begin{itemize}[noitemsep,topsep=0pt]
    \item Default SSL augmentations (\textit{RandomResizedCrop, RandomRotate, RandomHorizontalFlip, RandomVerticalFlip, RandomColorJittering, RandomGrayscale, GaussianBlur at various probabilities, $p$})
    \item SGD ($p=1.0$),
    \item SGD ($p=1.0$) with default SSL augmentations
    \item SGD ($p=0.5$) with default SSL augmentations
    \item SGD ($p=1.0$) with jittering ($p=1.0$)
    \item SGD ($p=0.5$) with jittering ($p=1.0$)
    \item SGD ($p=0.8$) with jittering ($p=0.8$)
\end{itemize}

The batch size is known to be one of the most important parameters in SSL and the larger the batch size, the better the representations as most of SSL algorithms greatly benefit from large batches (in order to have larger number of negatives available in mini batches) but we could not investigate performance for more than 128 images in a batch due to limited computational resources and the high number of experiments that we need to run (more than 500 runs, each takes at least a few hours).
%\tb{you are not playing; say "investigate performance for more than 128 images in a batch", or the like.} \vk{fixed}
In order to test the impact of image resizing and batch sizing on SGD in image segmentation tasks, we picked three parameter combinations with five SSL methods (SimSiam, SimCLR, BYOL, MoCo, Barlow Twins): (i) Offline SGD applied on the original size ($100p$) with batch size 16 ($bs16$), (ii) Offline SGD applied on the original size ($100p$) with batch size 128, (iii) Offline SGD applied on the upscaled size ($200p$) with batch size 16. 

Upscaling (x2) is tested to see the impact of detailed SGD segmentations, and it is found that the performance reaches its peak when SGD is applied on the original image-size with a batch size of 128. %and it is observed that the performance of augmentation strategies vary by SSL methods accompanied. Using original image sizes with batch size 128, the best performing strategies for SGD based ones are BYOL and SimSiam while the best ones for default augmentation strategy is Barlow and MoCo as can be seen at Figure ~\ref{figure:sgd_vs_aug_selection_per_bs}. 
We also tested with doubling the resolution of each image and then applying SGD but the impact of SGD on SSL performance was not as good as in the original image-size. 
This can be explained by the fact that the color variation in SGD with low resolution (original size) images is more or less similar to what we can expect to see in a solar panel segmentation. Once we increase the resolution by up-scaling, the details on a solar panel become more evident and the SGD may assign different colors for such areas, which we do not want to see for a simple panel segmentation task (e.g., every object is represented by a single color in image segmentation).

\begin{figure}[h]
  \includegraphics[width=1.0\columnwidth]{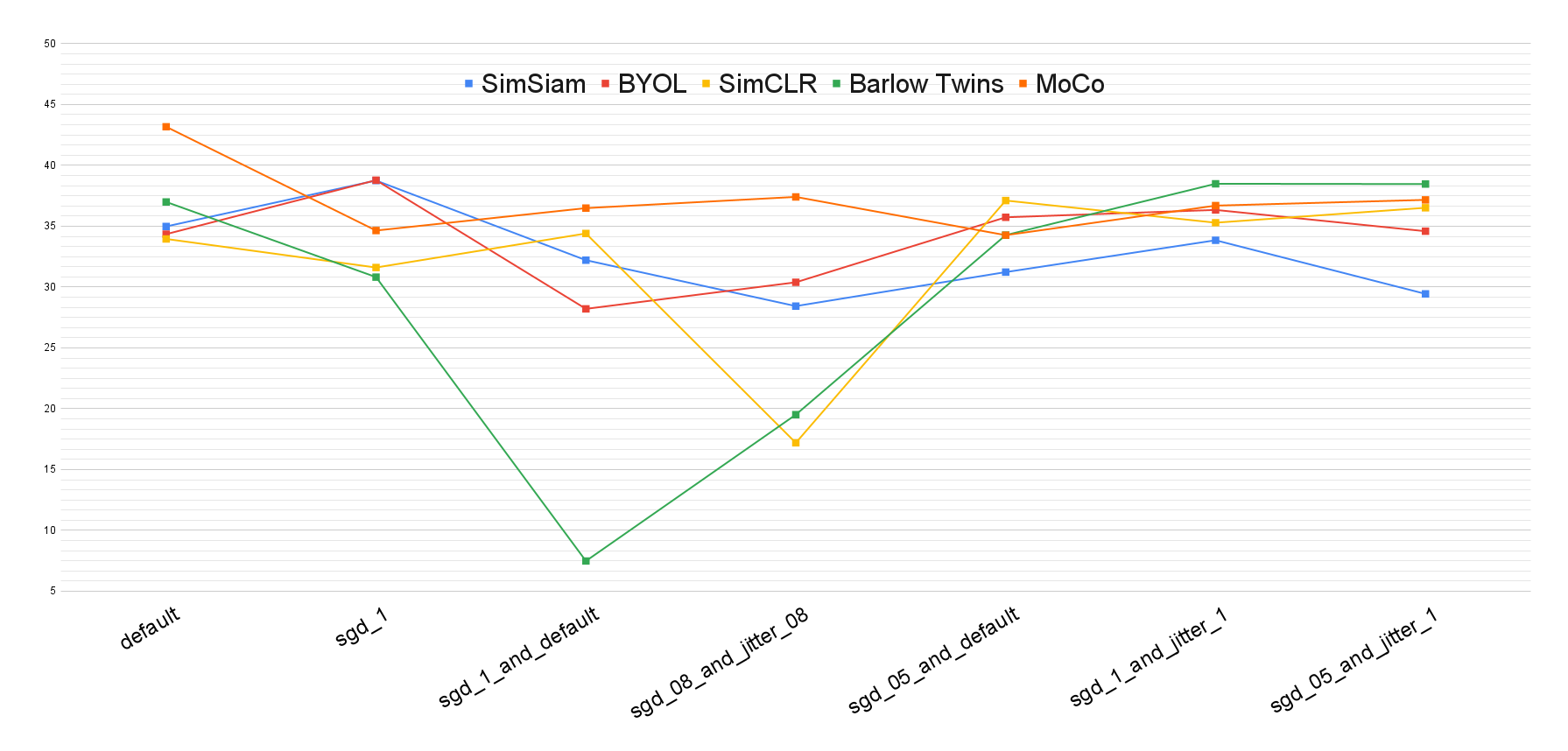}
\caption{Selecting the best combination of SGD and default augmentation strategies for various SSL methods using original size images with batch size 128. The performance of augmentation strategies varies by the SSL methods accompanied. SGD alone without any other augmentation strategies produced the best result in BYOL and SimSiam compared to other methods and augmentation policies. 
%The first row shows the average precision (AP) metrics for segmentations and second row shows the AP metric for object detection. In most of the SSL methods, the performance gain is the highest with SGD applied on original size images with batch size 128.~\vk{charts will be prettified -- same y axis and ranges}
}
\vspace{-5mm}
\label{figure:sgd_vs_aug_selection_per_bs}
\end{figure}

Then, using 500 images as a training set (out of 645 images from MultiRes-PV dataset), we pretrained all of the five SSL methods in a training loop with image segmentation and object detection tasks and measured the test accuracy on the test set (145 images). We observed that performances of SSL methods vary given the augmentation policy, as shown in Figure~\ref{figure:sgd_vs_aug_selection_per_bs}. 
For instance, while MoCo performs better compared to other SSL methods, with default SSL augmentations yet no SGD, SimSiam and BYOL perform better compared to other SSL methods with SGD plus default SSL augmentations as well as with only default augmentations. In another case, SimCLR performs the best when SGD is applied at $p=0.5$ along with default augmentations. This result tells us that each SSL method performs differently, as a function of the augmentation policy, while the impact of SGD also varies under different settings.
%\tb{This figure is crucial but cannot be read at all.} \vk{improved}

\begin{figure*}[h]
\includegraphics[width=1.0\textwidth]{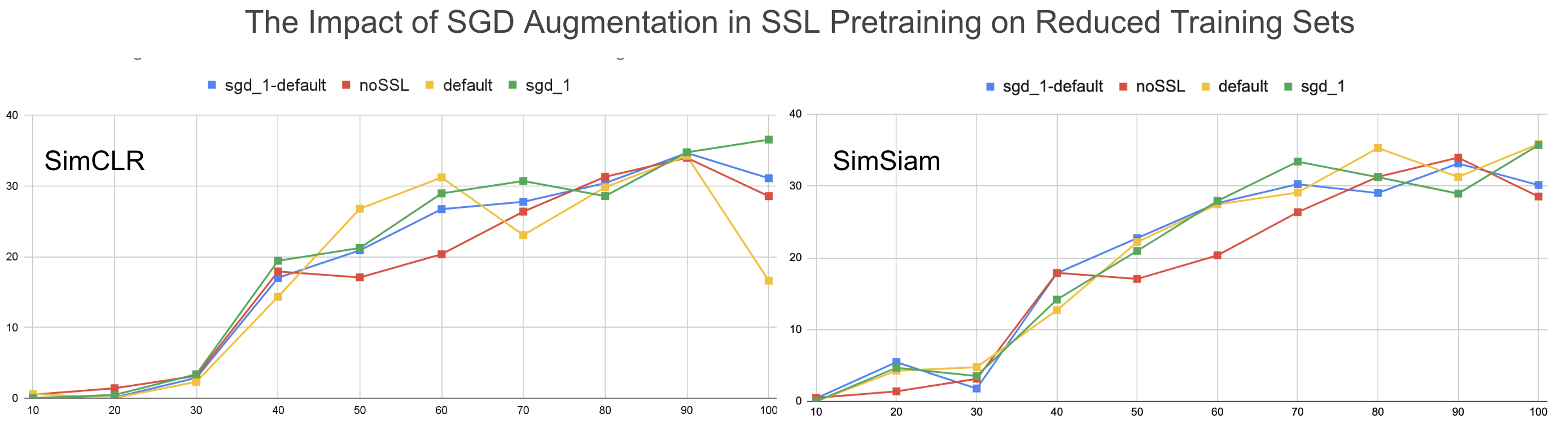}
\caption{Segmentation metrics for SGD with SimSiam and SimCLR experiments with reduced number of images from the training set at various levels. $x$-axis denotes the percentage of the samples with respect to entire training set (from 10\% to 100\%) and $y$-axis denotes test set average precision (AP) of segmentation model trained with the partial training set.}
\vspace{-5mm}
\label{figure:multires_sgd_partial_training}
\end{figure*}

Then, we experimented with the same SSL methods, having additionally no-SSL (with default ImageNet checkpoints)
at various levels of labeled images (subsets from 10\% to 100\%) and measured the test accuracy on the same 145 images. 
For instance, we picked at first 50 labeled images from the training set, then using a pretrained SSL backbone from the previous step, we trained an image segmentation and object detection model with Detectron2, and tested the models on the test set. Then we did the same for 20\%, 30\% and so on. The AP metrics for the selected SSL methods for selected augmentation are shown in Figure~\ref{figure:multires_sgd_partial_training}. 
As seen from this figure, in every level of the reduced training set, an augmentation policy with an SGD component usually performs better than an SSL with default augmentations as well as a vanilla (no-SSL) model training. In some cases, using only SGD augmentation would even suffice without further application of augmentation. 

%To accomplish this, we converted MultiRes-PV binary segmentation masks to COCO format and ...

\subsubsection{Image Classification Using SSL Backbones}

\begin{figure}[h]
  \centering
  \includegraphics[width=1.0\columnwidth]{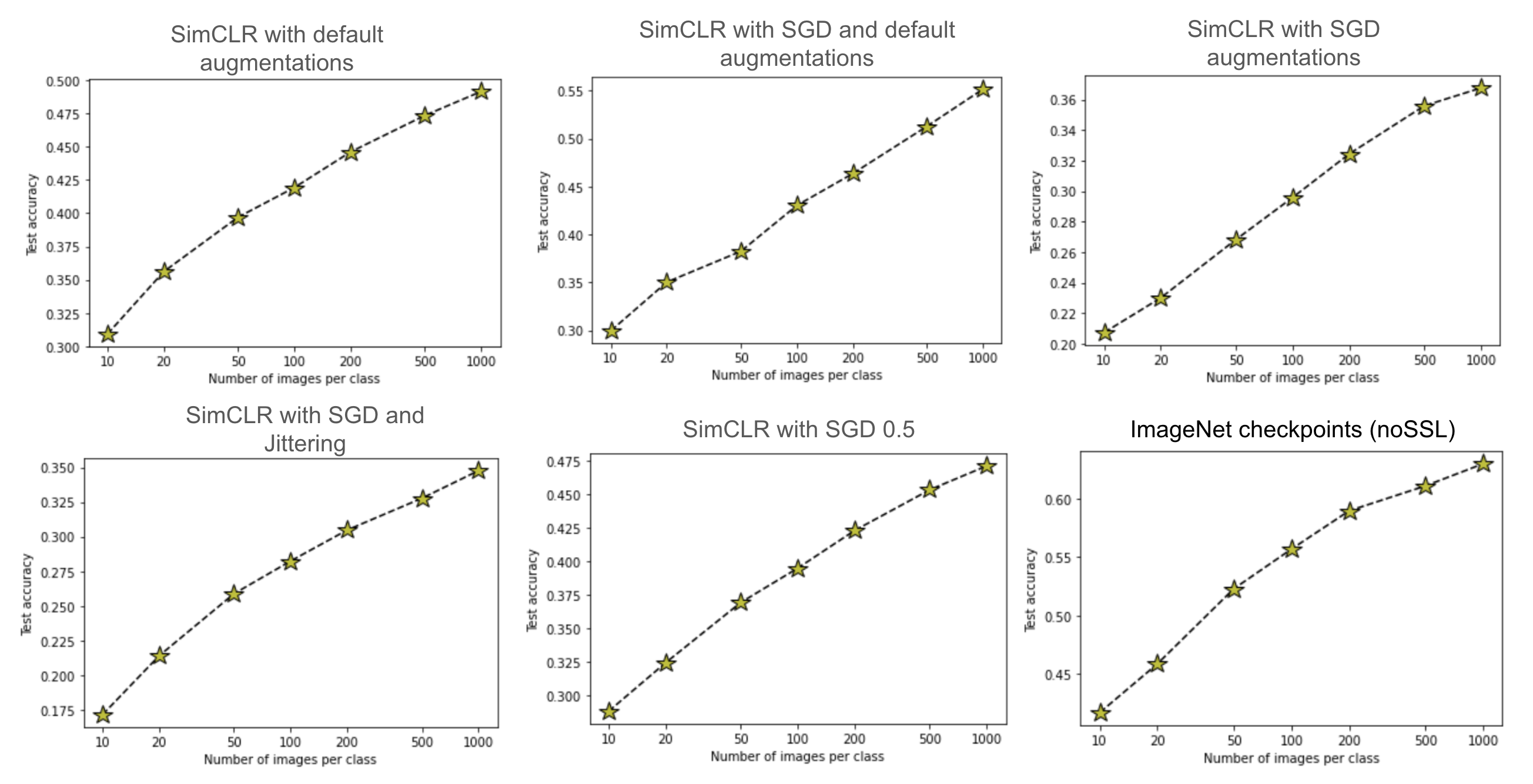}
\caption{Classification of images in CIFAR10 dataset using partial samples from training set and SSL method with SGD applied. SGD plus default augmentations outperforms default SSL augmentations by 5\% and a little worse than ImageNet checkpoints.}
\vspace{-5mm}
\label{figure:cifar10_SGD_logreg_clf}
\end{figure}

As a downstream image classification model, using CIFAR10 and STL10 datasets, we used Logistic Regression to classify the image representations taken from the penultimate layers of SimCLR backbones that are pretrained on CIFAR10 training set (5000 unlabelled images) with the augmentation policies mentioned before (with hashing applied to SGD augmentations). 
During these experiments, various levels of labeled images from CIFAR10 (from 10\% to 100\%) are picked at each iteration and only one SSL method (SimCLR) is picked for the sake of brevity. The results from this experiment can be seen in Figure~\ref{figure:cifar10_SGD_logreg_clf}). Even though there is a slim performance gain (around 5\% improvement) with SGD plus default augmentations applied, the default pretrained ImageNet backbone still performs better without SSL. This can be explained by the smaller number of utilized epochs (100) while pretraining with SimCLR and the large number of labelled images being used in the ImageNet pretraining. 
Considering that SimCLR is pretrained only by 5000 unlabelled images from CIFAR10 while the default ResNet18 model is pretrained with the entire ImageNet dataset with ground truth labels, getting similar metrics with SimCLR is still worth mentioning herein.

\begin{figure}[h]
  \centering
  \includegraphics[width=1.0\columnwidth]{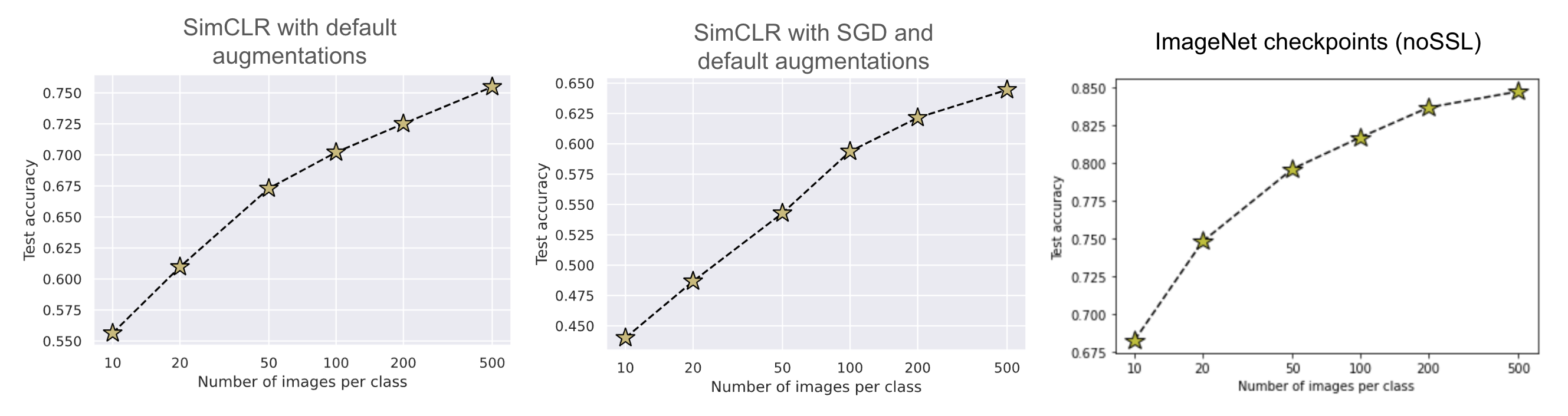}
\caption{Classification of images in STL10 dataset using partial samples from training set and SSL method with SGD applied. SGD did worse than default SSL augmentations and the default pretrained ImageNet backbone outperforms all the metrics achieved with SSL methods (due to the fact that STL is already a subset of ImageNet dataset).}
\vspace{-5mm}
\label{figure:STL10_SGD_logreg_clf}
\end{figure}

We run a similar experiment on the STL10 dataset (using 100 thousand unlabeled images), but SGD did not contribute in that case and performed worse. 
Since STL is already a subset of the ImageNet dataset, the default pretrained ImageNet backbone did quite well and exceed all the metrics achieved with SSL methods. The results are shown in Figure~\ref{figure:STL10_SGD_logreg_clf}.  Given the fact that the images in the STL10 dataset have higher resolution (96x96) than the images in CIFAR10 (32x32), getting worse performance with SSL using SGD can be explained by the same phenomenon we observed in SGD of upscaled images doing worse compared to original size images. %Ignoring ImageNet result due to this fact,

\section{Discussion}
\label{sec:discussion}

Our experiments with clustering the image representations via SGD-augmented backbone networks indicate that SGD enables obtaining better image representations in all of the cases that we tested. 
Even if we may not know the number of clusters within a dataset, the clear delineation between the clusters indicates a better representation to separate one image from another using visual clues. 
Usually, the most salient details and objects play the role of separators between one image from another, but using classical image augmentation policies (e.g., cropping, jittering, etc.) totally fails on this end. 
At the same time, SGD-based salient object segmentation does a better job on catching those critical details, thus assisting more in downstream clustering tasks. 

We argue that using a salient image segmentation in SSL generates better representations when the downstream task is image segmentation or object detection due to the fact that salient object detection and segmentation approaches segment only the salient foreground object from the background, rather than partition an image into regions of coherent properties as in general segmentation algorithms. Our experiments with various augmentation combinations with and without SGD show that the augmentation policy having an SGD component usually performs better than an SSL with default augmentations as well as a vanilla (no SSL) model training. 
In some cases, using only SGD augmentation would even suffice, with no further application of augmentation. 
 
Another important observation is that  each SSL method performs differently as a function of the underlying augmentation policy, whereas the impact of SGD also varies under different settings. For instance, while MoCo performs better compared to other SSL methods, with default SSL augmentations yet no SGD, SimSiam and BYOL perform better compared to other SSL methods with SGD plus default SSL augmentations as well as with only default augmentations, as was elaborated in the previous section. 
Notably, this in fact the common question in the age of DL in which there are a myriad of alternatives: How to know which algorithm does better on a certain use case and how to pick the right one? Not only the SSL method but also the augmentation technique, downstream task and even the resolution of your images matter.

One of the most unexpected observation of our tests can be regarded as having worse results with SGD applied on high resolution images. We have observed this effect on various occasions during our experiments as reported in the previous section. 
This can be explained by the fact that the color variation in SGD with low resolution (original size) images is lower and more similar to what we can expect to see in a coarse grained segmentation tasks. Once we increase the resolution by up-scaling, the details on an image becomes more evident and SGD may assign different colors for such areas, which we do not want to see for a simple image segmentation task (e.g., every object is represented by a single color in image segmentation). However, SSL performing well on low resolution images can also be attributed to the low number of epochs (i.e., 100) as the model is not able to catch the low level details in high resolution images in such a limited number of epochs. 

Nevertheless, a recent SSL study also supports our claim and sheds some light on this issue as follows:
\cite{cole2022does} claims that current self-supervised methods learn representations that can easily disambiguate coarse-grained visual concepts like those in ImageNet. However, as the granularity of the concepts becomes finer, self-supervised performance lags further behind supervised baselines. 

\section{Conclusions}\label{sec:conclusion}
By empirically addressing the posed research questions, our investigation confirmed the capacity of SGD to play a role in modern SSL.
Overall, this study successfully leveraged SGD, a 10-years-old salient object detection and segmentation algorithm, as a potent image augmentation technique for downstream image segmentation tasks. 
To achieve an effective integration of SGD into SSL pretraining routines, we devised a simple manipulation called offline augmentation with hashing. 
This fine implementation detail enabled us to run hundreds of SSL experiments with various parameters and configurations. 
We then demonstrated that using a salient image segmentation in SSL generates better representations when the downstream task is image segmentation. Our experiments with clustering the image representations via SGD-augmented backbone networks indicate that SGD helps better image representations in all of the cases tested. Our experiments with various augmentation policies including SGD show that the augmentation policy having a SGD component usually does better than a SSL with default augmentations; in some cases, using only SGD augmentation alone would even be be better. 

Our experiments with MultiRes-PV, CIFAR10 and STL10 also showed that SSL with SGD-based augmentation policy performs well with low resolution images. This still remains to be verified whether fine-grained features and/or the low number of epochs played a role in this observation. 
We contend that salient object segmentation algorithms produce coarse grained segmentation due to saliency, thus perform well on segmenting coarse grained objects like PV solar panels and it may not hold true if our goal were to extract fine grained details in an image.
 
We observed that each SSL method performs differently given the augmentation policy, whereas the impact of SGD also varies under different settings.  In our opinion, the most unexpected observation of our tests can be regarded as having worse results with SGD applied to high resolution images compared to low resolution ones. 
We conclude that the augmentation technique, type of a downstream task and image resolution are the most important elements that have a high impact on the success of a SSL method picked.

As a future work, harnessing the offline augmentation with hashing, we plan to investigate if the observations gathered during this study would still hold true with any other unsupervised or zero-shot segmentation as well as salient object detection algorithms that are used as an augmentation policy in SSL pretraining process.
%\clearpage
\bibliographystyle{unsrtnat}
\bibliography{References}

\end{document}